\pdfoutput=1

\documentclass[11pt]{article}

\usepackage[final]{acl}

\usepackage{times}
\usepackage{latexsym}

\usepackage[T1]{fontenc}

\usepackage[utf8]{inputenc}

\usepackage{microtype}

\usepackage{inconsolata}

\usepackage{hyperref}
\usepackage{url}
\usepackage{booktabs}
\usepackage{enumerate}
\usepackage{algorithm}
\usepackage{algorithmic}
\usepackage{multirow}

\usepackage{amsmath}
\usepackage{amssymb}
\usepackage{mathtools}
\usepackage{amsthm}
\usepackage{bm}
\usepackage{enumerate}
\usepackage{booktabs}

\usepackage[absolute,overlay]{textpos}

\title{Complex Logical Query Answering by Calibrating Knowledge Graph Completion Models}

\author{Changyi Xiao\textsuperscript{\rm 1}\thanks{Corresponding author.},
    Yixin Cao\textsuperscript{\rm 2} \\
  \textsuperscript{\rm 1}School of Data Science, University of Science and Technology of China\\
  \textsuperscript{\rm 2}School of Computer Science, Fudan University\\
  \texttt{changyi@mail.ustc.edu.cn, caoyixin2011@gmail.com}\\}

\begin{document}
\maketitle
\begin{abstract}
Complex logical query answering (CLQA) is a challenging task that involves finding answer entities for complex logical queries over incomplete knowledge graphs (KGs). Previous research has explored the use of pre-trained knowledge graph completion (KGC) models, which can predict the missing facts in KGs, to answer complex logical queries. However, KGC models are typically evaluated using ranking evaluation metrics, which may result in values of predictions of KGC models that are not well-calibrated. In this paper, we propose a method for calibrating KGC models, namely CKGC, which enables KGC models to adapt to answering complex logical queries. Notably, CKGC is lightweight and effective. The adaptation function is simple, allowing the model to quickly converge during the adaptation process. The core concept of CKGC is to map the values of predictions of KGC models to the range [0, 1], ensuring that values associated with true facts are close to 1, while values linked to false facts are close to 0. Through experiments on three benchmark datasets, we demonstrate that our proposed calibration method can significantly boost model performance in the CLQA task. Moreover, our approach can enhance the performance of CLQA while preserving the ranking evaluation metrics of KGC models. The code is available at \url{https://github.com/changyi7231/CKGC}.
\end{abstract}

\begin{textblock*}{\textwidth}(25mm,282mm)
    \centering
    {\small \textit{Findings of the Association for Computational Linguistics ACL 2024, pages 13792–13803}\\
    August 11-16, 2024 ©2024 Association for Computational Linguistics
    }
\end{textblock*}

\section{Introduction}
\begin{figure*}[t]
\begin{center}
\centerline{\includegraphics[width=1.0\textwidth]{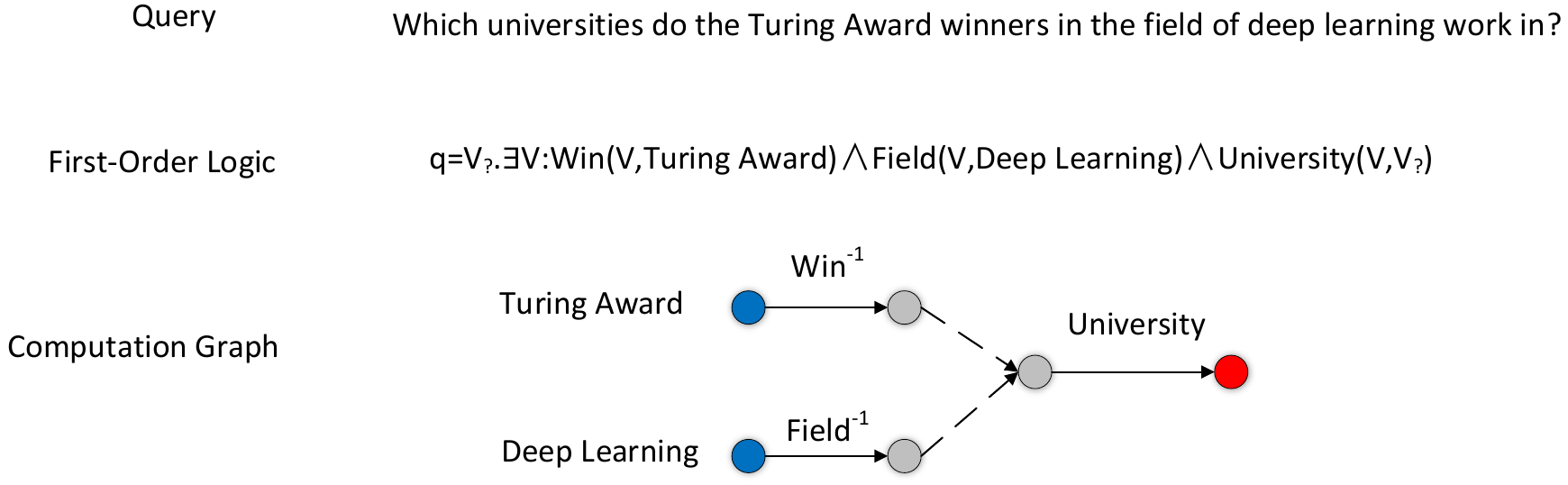}}
\caption{An query, its corresponding FOL form and its corresponding computation graph.}
\label{figure:1}
\end{center}
\end{figure*}

Knowledge graphs (KGs) are composed of structured representations of facts in the form of triplets and have been widely used in various domains. One of the key tasks associated with KGs is complex logical query answering \cite{ren2023neural}. Complex logical queries are typically expressed using first-order logic (FOL), which encompasses logical operations such as conjunction ($\wedge$), disjunction ($\vee$), negation ($\neg$), and existential quantifier ($\exists$). For example, the query "Which universities do the Turing Award winners not in the field of deep learning work in?" can be formulated as a FOL query, as illustrated in Figure \ref{figure:1}.

Many well-known knowledge graphs (KGs) suffer from incompleteness, rendering it challenging to answer complex queries through simple KG traversal. Building on the accomplishments of knowledge graph completion (KGC) methods \cite{bordes2013translating, sun2018rotate, trouillon2017knowledge} in addressing one-hop KG queries, a research avenue has emerged focusing on learning embeddings for queries to handle complex logical queries \cite{hamilton2018embedding, ren2020query2box, zhang2021cone}. Nonetheless, these methodologies often require extensive training on numerous complex logical queries, leading to substantial training time overhead and limited generalization to out-of-distribution query structures.

In addressing these challenges, CQD \cite{arakelyan2020complex} introduces a method for CLQA by leveraging one-hop atom results derived from a pretrained KGC model, thereby removing the necessity for training on complex queries. CQD frames CLQA as an optimization problem and employs techniques like beam search or continuous approximation to estimate the optimal solution. Despite its effectiveness, the approximations made during the process could lead to a decrease in accuracy for CQD. Furthermore, CQD is reliant on a KGC model whose values of output might not be specifically calibrated for CLQA, potentially resulting in inaccuracies in the outcomes.

In this paper, we introduce a method for calibrating KGC models, namely CKGC, which can make KGC models adapt to handling complex logical queries. Our method is lightweight and effective. It is lightweight, as the adaptation function is simple and the model can quickly converge in adaptation process. Moreover, it is effective in significantly enhancing the performance of CLQA.

We first represent every complex logical query as a computation graph over fuzzy sets, of which nodes represent fuzzy sets and edges represent the fuzzy set operations over fuzzy sets. To obtain the answers of a query, we traverse the computation graph and execute the fuzzy set operations through a straightforward forward propagation process, eliminating the need for optimization compared to CQD \cite{arakelyan2020complex}. We define four fundamental fuzzy set operations for CLQA: projection operation, complement operation, intersection operation, and union operation, all of which require the KGC model to be calibrated.

However, a KGC model is typically measured by the ranking metrics, which do not need it to be calibrated. Thus, we propose a calibration method for KGC models to make it adapted to CLQA. We define the adaptation function as a monotonically increasing function that preserves the ranking evaluation metrics of the KGC model. The main idea of our method is to map the values of output of a KGC model to the interval $[0,1]$, and makes the values corresponding to true triplets be as close as possible to 1, while the values corresponding to false triplets be as close as possible to 0.

Our experimental evaluation on three standard datasets demonstrates the efficacy of CKGC in CLQA. The results reveal that CKGC achieves state-of-the-art performance across three datasets, showcasing an average relative improvement of 6.7$\%$ on existential positive first-order queries and 53.9$\%$ on negation queries compared to the prior state-of-the-art method.

\section{Related Work}
\paragraph{Knowledge Graph Completion}
The task of KGC involves predicting missing triplets within a knowledge graph, which can be viewed as predicting answers for one-hop queries. Various methodologies have been proposed to address this task, encompassing embedding techniques \citep{bordes2013translating, sun2018rotate, trouillon2017knowledge}, reinforcement learning approaches \cite{xiong2017deeppath, das2018go, hildebrandt2020reasoning, zhang2022learning}, rule learning strategies \cite{yang2017differentiable, sadeghian2019drum, qu2020rnnlogic}, and graph neural network methodologies \cite{schlichtkrull2018modeling, vashishth2019composition, teru2020inductive}. Embedding methods aim to embed entities and relations into a continuous space and define a scoring function based on these embeddings. Reinforcement learning methods train an agent to explore the knowledge graph to predict the missing triplets. Rule learning methods adopt a distinct approach by initially identifying confident logical rules from the knowledge graph. These rules are then utilized to infer missing triplets. Lastly, graph neural network methods utilize graph neural networks to learn representations of entities and relations by leveraging the graph structure.

\paragraph{Complex Logical Query Answering}
CLQA over KGs extends KGC to predict answers for FOL queries, which additionally requires defining relationships between sets of entities. Embedding-based methods represent sets of entities as geometric objects \cite{hamilton2018embedding,ren2020query2box,zhang2021cone} or probability distributions \cite{ren2020beta}, and then minimize the distance between embeddings of queries and embeddings of their corresponding answers. However, the quality of representation of sets may be compromised when dealing with large sets. To overcome this limitation, some studies have incorporated powerful fuzzy set theory to handle FOL queries \cite{chen2022fuzzy,zhu2022neural}. For instance, FuzzQE \cite{chen2022fuzzy} embeds entities and queries into a fuzzy space and leverages fuzzy set operations to perform logical operations on the embeddings. Similarly, GNN-QE \cite{zhu2022neural} decomposes the query into relational projections operations and fuzzy set operations over fuzzy sets, and subsequently learns a graph neural network to execute relational projections.

Nevertheless, the aforementioned methods generally necessitate training on numerous complex logical queries, resulting in significant training time overhead and limited generalization to out-of-distribution query structures. Another line of methods first pre-trains a KGC model and then integrates it with fuzzy set theory to infer answers. CQD \cite{arakelyan2020complex} formulates CLQA as an optimization problem and proposes two strategies to approximate the optimal solution: CQD-CO, which directly optimizes in the continuous space, and CQD-Beam, which utilizes beam search. Despite CQD avoiding the need to train on complex logical queries, it suffers from accuracy loss due to the approximated optimization and uncalibrated KGC models. To enhance accuracy, \citet{bai2023answering} introduce QTO, which efficiently finds the theoretically optimal solution through a forward-backward propagation on a tree-like computation graph. Additionally, \citet{arakelyan2023adapting} propose $\text{CQD}^{\mathcal{A}}$, a parameter-efficient adaptation method, to calibrate KGC models. However, the KGC model for QTO is not well calibrated for complex queries, and the optimal solution for $\text{CQD}^{\mathcal{A}}$ is still approximated. In contrast, our proposed model not only delivers accurate solutions but also effectively calibrates the KGC models.

\section{Background}
In this section, we introduce the related background of our method, complex logical query answering on knowledge graphs and fuzzy sets.
\subsection{Complex Logical Query Answering on Knowledge Graphs}
\label{section:3.1}
\paragraph{Knowledge Graphs Completion}
Given a set of entities $\mathcal{V}$ and a set of relations $\mathcal{R}$, a knowledge graph $\mathcal{G}$ contains a set of triplets $\{(h,r,t)\}\subset \mathcal{V}\times \mathcal{R}\times \mathcal{V}$, where each triplet is a fact from head entity $h$ to tail entity $t$ with a relation type $r$. KGC models define scoring functions $f(h,r,t)$ to measure the likelihood of triplets $(h,r,t)$ based on their corresponding embeddings.

\paragraph{First-Order Logic Queries}
Complex logical queries can be represented by FOL queries with logical operations including conjunction ($\wedge$), disjunction ($\vee$), negation ($\neg$), and existential quantifier ($\exists$). A first-order logic query $q$ can be described in its disjunctive normal form, which consists of a set of non-variable anchor entities $\mathcal{V}_{a}\subseteq \mathcal{V}$, existentially quantified bound variables $V_{1},\dots,V_{k}$ and a single target variable $V_{?}$, which provides the answers of query $q$. The disjunctive normal form of a logical query $q$ is a disjunction of one or more conjunctions.
$$q=V_{j}.\exists V_{1},\dots,V_{k}:c_{1}\wedge c_{2}\wedge,\cdots,c_{n}.$$
Each $c$ represents a conjunctive query with one or more literals $e$, i.e., $c_{i}=e_{i1}\vee e_{i2}\vee \cdots \vee e_{im}$. Each literal $e$ represents an atomic formula or its negation, i.e.,  $e_{ij}=R(v_{a},V)$ or $\neg R(v_{a},V)$ or $\neg R(V^{'},V)$ or $\neg R(V^{'},V)$, where $v_{a}\in \mathcal{V}_{a}$, $V\in \{V_{?},V_{1},\dots,V_{k}\}$, $V\in \{V_{?},V_{1},\dots,V_{k}\}$, $V\neq V^{'}$, $R(\cdot, \cdot)$ is a binary function $R: \mathcal{V}\times \mathcal{V}\xrightarrow{}\{0,1\}$. Each relation $r\in \mathcal{R}$ corresponds to a binary function $R(\cdot, \cdot) $. If $(h,r,t) $is a true fact, then $R(h, t)=1 $. If $(h,r,t) $is a false fact, then $R(h, t)=0$.

\paragraph{Computation Graph}
Given a FOL query, we can represent it as a computation graph, of which nodes represent sets of entities and edges represent set operations over sets of entities. The set operations include the complement operation, intersection operation, union operation and projection operation. The root node represent the set of answer entities. See Figure \ref{figure:1} for an example. We map logical operations to set operations according to the following rules.
\begin{itemize}
    \item \textbf{Negation}$\longrightarrow{}$\textbf{Complement Operation:} Given a set of entities $S\subseteq \mathcal{V}$, the complement operator performs set complement to obtain $\overline{S}=\mathcal{V}\backslash S$.
    \item \textbf{Conjunction}$\longrightarrow{}$\textbf{Intersection Operation:} Given $n$ sets of entities $\{S_{1},S_{2},\dots S_{n}\}$, the intersection operator performs set intersection to obtain $\cap_{i=1}^{n}S_{i}$.
    \item \textbf{Disjunction}$\longrightarrow{}$\textbf{Union Operation:} Given $n$ sets of entities $\{S_{1},S_{2},\dots S_{n}\}$, the union operator performs set union to obtain $\cup_{i=1}^{n}S_{i}$.
    \item \textbf{Relation Projection}$\longrightarrow{}$\textbf{Projection Operation:} Given a set of entities $S\subseteq \mathcal{V}$ and a relation $r\in \mathcal{R}$,  the projection operator outputs all the adjacent entities $\cup_{v\in S}N(v,r)$, where $N(v,r)$ is the set of entities such that $(v,r,v^{'})$ are true triplets for all $v^{'}\in N(v,r)$.
\end{itemize}
In order to answer a given FOL query, we can traverse the computation graph and execute the set operations. The answers of a query can be obtained by looking at the set of entities in the root node.

\subsection{Fuzzy sets}
\paragraph{Definition}
A fuzzy set is a pair $(U,m)$, where $U$ is a set and $m:U\xrightarrow{}[0,1]$ is a membership function. For each $x\in U$, the value $m(x)$ measures the degree of membership of $x$ in $(U,m)$. The function $m=\mu _{A}$ is called the membership function of the fuzzy set $A=(U,m)$. Classical sets are be seen as special cases of fuzzy sets, if the membership functions only takes values 0 or 1.

\paragraph{Fuzzy Set Operations}
Fuzzy set operations are a generalization of classical set operations for fuzzy sets. The three primary fuzzy set operations are fuzzy complements, fuzzy intersections, and fuzzy unions. For a given fuzzy set $A$, its complement $\overline{A}$ is commonly defined by the following membership function:
$$\forall x\in U, \mu_{\overline{A}}(x)=1-\mu_{A}(x)$$
Given a pair of fuzzy sets $A,B$, their intersection $A\cap B$ is defined by a t-norm $T$ \cite{klement2004triangular}:
$$\forall x\in U, \mu_{A\cap B}(x)=T(\mu_{A}(x),\mu_{B}(x))$$
Prominent examples of t-norms include product t-norm $T(a,b)=ab$, Gödel t-norm $T(a,b)=\min\{a,b\}$ and so on \cite{klement2004triangular}.
Given a t-norm $T$ and a pair of fuzzy sets $A,B$, their union $A\cup B$ is defined by De Morgan's law:
\begin{align*}
    \forall x\in U, \mu_{A\cup B}(x)=1-T(1-\mu_{A}(x),1-\mu_{B}(x))
\end{align*}

\section{Method}
We first define four fuzzy set operations over fuzzy sets in Section \ref{section:4.1}, and then propose a calibration method for KGC models in Section \ref{section:4.2}.
\subsection{Fuzzy Set Operations}
\label{section:4.1}
As demonstrated in Section \ref{section:3.1}, a query can be represented as a computation graph. To obtain the fuzzy sets of answers, it is necessary to define the fuzzy set operations utilized in computation graphs. These operations encompass the projection operation, intersection operation, union operation, and complement operation.

We represent every fuzzy set of entities as a vector $\bm{e}\in [0,1]^d$, where $d=|\mathcal{V}|$ and $\bm{e}_{i}$ denotes the grade of membership of entity $i$. The anchor entity is represented by a vector with a single element set to 1 and all other elements set to 0.

\paragraph{Complement Operation:}
The complement operation maps a fuzzy set to the complement of the fuzzy set. We define the complement operation as
\begin{align*}
\mathcal{C}:[0,1]^{d}\xrightarrow{}&[0,1]^{d}\\
\mathcal{C}(\bm{e})=&\bm{1}-\bm{e}
\end{align*}

\paragraph{Intersection Operation:}
The intersection operation maps several fuzzy sets into the intersection set of these fuzzy sets. We define the intersection operation by utilizing the product t-norm, which is as follows:
\begin{align*}
    \mathcal{I}:[0,1]^{nd}\xrightarrow{}&[0,1]^{d}\\
\mathcal{I}(\bm{e}^{1},\bm{e}^{2},\dots,\bm{e}^{n})=&\bm{e}^{1}\odot \bm{e}^{2}\odot \dots\odot \bm{e}^{n}
\end{align*}
where $\{\bm{e}^{i}\in [0,1]^d|1\leq i\leq n\}$ are vector representations of fuzzy sets and $\odot$ is the Hadamard product.

\paragraph{Union Operation:}
The union operation maps several fuzzy sets into the union set of these fuzzy sets. Due to the De Morgan's Law, we do not need to define the union operation directly. We define the union operation by utilizing the intersection operation and complement operation as follows:
\begin{align*}
\mathcal{U}:&[0,1]^{nd}\xrightarrow{}[0,1]^{d}\\
\mathcal{U}(\bm{e}^{1},\bm{e}^{2},\dots,\bm{e}^{n})=&\mathcal{C}(\mathcal{I}(\mathcal{C}(\bm{e}^{1}),\mathcal{C}(\bm{e}^{2}),\dots,\mathcal{C}(\bm{e}^{n})))
\end{align*}

\paragraph{Projection Operation:}
The projection operation maps a fuzzy set into another fuzzy set with a relation type $r$. We implement the projection operation by utilizing the Gödel t-norm as follows:
\begin{align*}
\mathcal{P}_{r}:&[0,1]^{d}\xrightarrow{}[0,1]^{d}\\
\mathcal{P}_{r}(\bm{e})_{j}=&1-\min_{1\leq i\leq d}(1-\bm{e}_{i}\bm{X}_{i,r,j})\\
=&\max_{1\leq i\leq d}\bm{e}_{i}\bm{X}_{i,r,j}
\end{align*}
where $\bm{X}\in [0,1]^{|\mathcal{V}|\times |\mathcal{R}|\times |\mathcal{V}|}$ is the KG tensor and $\bm{X}_{i,r,j}$ denotes the likelihood of a triplet $(i,r,j)$, which is provided by the KGC model. We do not use product t-norm here because continued product often leads to accumulated errors for large $d$. Compared to QTO \cite{bai2023answering}, we define the projection operation and the complement operation separately, which make our method more flexible.

In the training process, since the value of $\bm{X}_{i,r,j}$ is constantly changed, we need to recompute the value of $\bm{X}_{i,r,j}$. The primary bottleneck of the projection operation is the computation. We can take the nonzero entries of $\bm{e}$ and multiply them with the corresponding entries in $\bm{X}$ to get results. Thus, the computational complexity of projection operation is $\mathcal{O}(|\mathcal{V}||\{\bm{e}_{i}|\bm{e}_{i}>0\}|)$.

In the test process, since the value of $\bm{X}_{i,r,j}$ is fixed, we pre-compute the value of $\bm{X}_{i,r,j}$ to reduce the computation. The primary bottleneck of the projection operation is the memory consumption of tensor $\bm{X}$. $\bm{X}$contains $|\mathcal{V}|^2|\mathcal{R}|$ entries. Due to the sparsity of the KG, most entries of $\bm{X}$ have small values, which can be filtered to 0 by a threshold $\epsilon>0$ while maintaining precision \cite{bai2023answering}.

\begin{figure*}[t]
\begin{center}
\centerline{\includegraphics[width=1.0\textwidth]{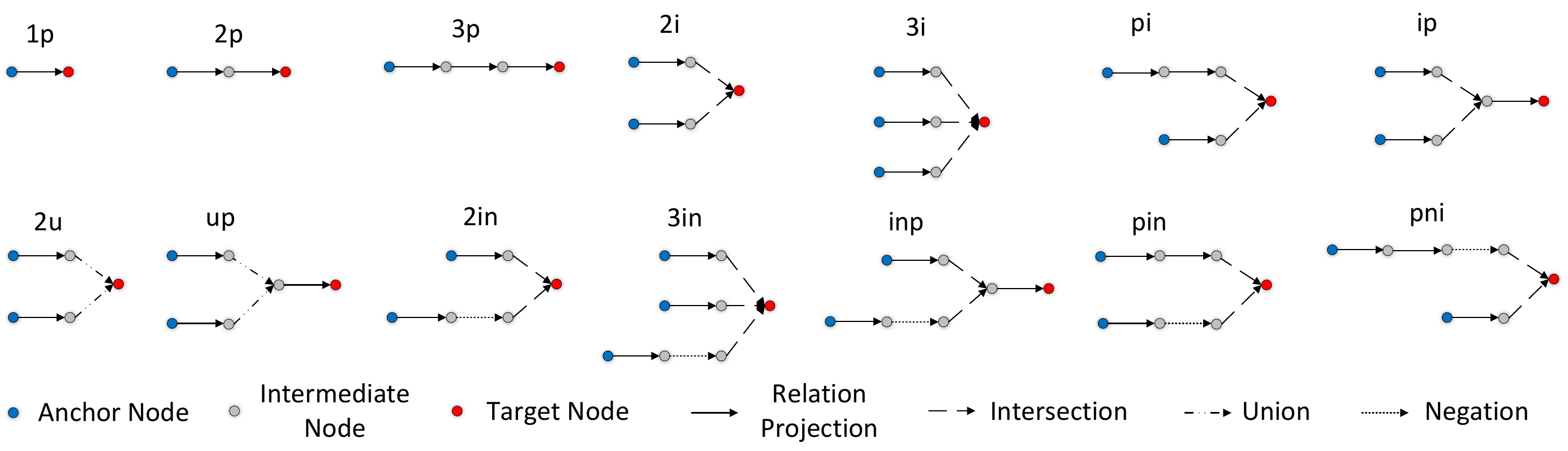}}
\caption{Fourteen types of queries used in the datasets, where "p" denotes relation projection, "i" denotes intersection, "u" denotes union, and "n" denotes negation.}
\label{figure:2}
\end{center}
\end{figure*}

\subsection{Calibration}
\label{section:4.2}
Upon establishing the four fuzzy set operations, the sole requirement for computing the results of queries is a calibrated KG tensor $\bm{X}$. The entries of a calibrated KG tensor $\bm{X}$ are expected to fall within the range of 0 to 1, with values associated with true triplets approaching 1, and those linked to false triplets nearing 0. In the event of possessing a fully calibrated KG tensor $\bm{X}$, accurate answers of queries can be attained through the utilization of the four fuzzy set operations.

The KG tensor $\bm{X}$ is furnished by a KGC model, which defines a scoring function $f(h,r,t)$ to measure the likelihood of a triplet $(h,r,t)$. The ranking metrics, MRR and H@N \citep{bordes2013translating}, are commonly used to evaluate KGC models. For example, the definition of MRR is as follows:
\begin{align*}
\text{MRR}=\sum_{(h,r,t)\in \mathcal{G}}\frac{1}{|\mathcal{G}|}\frac{1}{\text{rank}(h,r,t)}
\end{align*}
where $\mathcal{G}$ is a dataset and $\text{rank}(h,r,t)$ is the rank of tail entity $t$ in the predicted list for the query $(h,r,?)$. $\text{rank}(h,r,t)$ is computed based on the scoring function $f(h,r,t)$. The ranking metrics mainly focus on ranking rather than the specific numerical value of $f(h,r,t)$, which is important for complex query answering. Subsequently, an illustrative example is provided to elucidate this concept.

For example, supposing we have a query $$q=V_{?}:R_{1}(a_{1},V_{?})\wedge R_{2}(a_{2},V_{?})$$
where entity $a_{1}$ can be represented by a vector $\bm{a}_{1}=[1,0,0,0]$, entity $a_{2}$ can be represented by a vector $\bm{a}_{2}=[0,1,0,0]$. Assuming the KG tensor $\bm{X}$ is satisfied with $\bm{X}_{1,1,:}=[0.6, 0.4, 0.2, 0.1]$ and $\bm{X}_{2,2,:}=[0.5, 0.7, 0.2, 0.1]$. Then, we have that $\mathcal{P}_{1}(\bm{a}_{1})=\bm{X}_{1,1,:}=[0.6, 0.4, 0.2, 0.1]$, $\mathcal{P}_{2}(\bm{a}_{2})=\bm{X}_{2,2,:}=[0.5, 0.7, 0.2, 0.1]$ and the predicted answers 
$$\mathcal{I}(\mathcal{P}_{1}(\bm{a}_{1}),\mathcal{P}_{2}(\bm{a}_{2}))=[0.30, 0.28, 0.04, 0.01]$$
Then the largest entry in $\mathcal{I}(\mathcal{P}_{1}(\bm{a}_{1}),\mathcal{P}_{2}(\bm{a}_{2}))$ is the first entry.

If we let $\hat{\bm{X}}_{1,1,:}=0.1\bm{X}_{1,1,:}+0.1$, then we have that $\mathcal{P}_{1}(\bm{a}_{1})=\hat{\bm{X}}_{1,1,:}=[0.16, 0.14, 0.12, 0.11]$, $\mathcal{P}_{2}(\bm{a}_{2})=\bm{X}_{2,2,:}=[0.5, 0.7, 0.2, 0.1]$ and the predicted answers $$\mathcal{I}(\mathcal{P}_{1}(\bm{a}_{1}),\mathcal{P}_{2}(\bm{a}_{2}))=[0.08, 0.098, 0.024, 0.011]$$
Then the largest entry in $\mathcal{I}(\mathcal{P}_{1}(\bm{a}_{1}),\mathcal{P}_{2}(\bm{a}_{2}))$ is the second entry. 

While the aforementioned transformation does not alter the ranking metrics of KGC models, it can affect the ranking metrics of CLQA models. Hence, it is imperative to calibrate the KGC models to acquire a calibrated KG tensor $\bm{X}$ for the CLQA task.

We get a calibrated KG tensor $\bm{X}$ in the following steps:
\begin{enumerate}
    \item We train a KGC model to get a scoring function $f(h,r,t)$.
    \item We use softmax function to normalize $f(h,r,t)$ and denote the new scoring function as $\hat{f}(h,r,t)$.
    \begin{align*}
    &\hat{f}(h,r,t)=\min(N\frac{\exp(f(h,r,t))}{\sum_{t^{'}=1}^{|\mathcal{V}|}\exp(f(h,r,t^{'}))},1)\\
        &N=\begin{cases}
        M, & M>0\\
        \alpha, & M=0
    \end{cases}\\
    &M=|\{(h^{'},r^{'},t^{'})\in \mathcal{G}_{\text{train}}|h^{'}=h,r^{'}=r\}|
    \end{align*}
    where $\mathcal{G}_{\text{train}}$ denotes the training set of a KG. and $\alpha\geq 0$ is a hyper-parameter.
    \item We adapt the KGC models to complex queries datasets. $$\Tilde{f}(h,r,t)=\min(\bm{W}_{h,r}\hat{f}(h,r,t),1)$$
    where $\bm{W}\in (0,+\inf)^{|\mathcal{V}|\times |\mathcal{R}|}$ is a parameter matrix. For a query-answer pair $(q,[\![q]\!])$, where $[\![q]\!]$ denotes the set of answers of query $q$, we optimize $\bm{W}$ by the following loss function:
    $$\mathcal{L}=-\frac{1}{|[\![q]\!]|}\sum_{i\in [\![q]\!]}\log \bm{a}_{i}-\frac{1}{|\overline{[\![q]\!]}|}\sum_{i\in \overline{[\![q]\!]}}\log (1-\bm{a}_{i})$$
    where $\bm{a}$ is the vector representation of predicted answers.
    \item We replace the values of $\tilde{f}(h,r,t)$ with 1 for triplets $(h,r,t)\in \mathcal{G}_{\text{train}}\cup \mathcal{G}_{\text{validation}}$ to get the calibrated tensor $\bm{X}$, i.e.,
    \begin{align*}
        &\bm{X}_{h,r,t}=\\
        &\begin{cases}
        1 & (h,r,t)\in \mathcal{G}_{\text{train}}\cup \mathcal{G}_{\text{validation}}\\
        \tilde{f}(h,r,t) &(h,r,t)\notin \mathcal{G}_{\text{train}}\cup \mathcal{G}_{\text{validation}}
    \end{cases}
    \end{align*}
     where $\mathcal{G}_{\text{validation}}$ denotes the validation set of a KG.
\end{enumerate}

The first step is to get a pre-train KGC model, which lays the foundation for subsequent steps. Although any KGC model is allowed, it is better to choose a KGC model with good performance.

The second step is to normalize the scoring function $f(h,r,t)$ provided by the first step, which can make the values of $\hat{f}(h,r,t)$ between 0 and 1, and the larger the value of $f(h,r,t)$, the larger the value of $\hat{f}(h,r,t)$. A good property of softmax function is that it is invariant under translation transformation, i.e., $\text{softmax}(x+c)=\text{softmax}(x)$ for any $c\in \mathbb{R}$. The meaning of $N$ is the number of answer tail entities for a query $(h,r,?)$. As there can be multiple answer tail entities for a query $(h,r,?)$, each of their corresponding predicted values should be close to 1. Thus, we multiply $\text{softmax}(f(h,r,t))$ by $N$ to make it close to 1. The operation of taking the minimum value is to prevent the value from exceeding 1. The hyper-parameter $\alpha$ is to prevent $\hat{f}(h,r,t)$ from becoming 0 for the case where $M=0$.

The third step adapts the KGC models to CLQA datasets. Since the adaptation function, linear function, is a monotonically increasing function, the new scoring function $\tilde{f}(h,r,t)$ have the same results of KGC ranking metrics as $f(h,r,t)$. Thus, our method can improve the performance of CLQA while maintaining the evaluation results of KGC. This step is lightweight because the adaptation function is simple and only the parameter matrix $\bm{W}$ is optimized.

The fourth step replaces the value of $\tilde{f}(h,r,t)$ with 1 for known true triplets to get a calibrated KG tensor $\bm{X}$. While evaluating KGC models, the known true triplets are not typically taken into account. Nonetheless, this step is crucial for CLQA as it ensures the calibration of the tensor $\bm{X}$ for better performance.

\begin{table*}[t]
\begin{center}
\begin{tiny}
\begin{tabular}{lllllllllllllllll}
\toprule
Models & $\text{avg}_{p}$& $\text{avg}_{n}$& 1p& 2p& 3p& 2i& 3i& pi& ip& 2u& up& 2in& 3in& inp& pin& pni\\
\midrule
\multicolumn{17}{c}{FB15k}\\
\midrule
\multicolumn{1}{l}{GQE}&  28.0& -& 54.6& 15.3& 10.8& 39.7& 51.4& 27.6& 19.1& 22.1& 11.6& -& -& -& -& -\\
\multicolumn{1}{l}{Q2B}&  38.0& -& 68.0& 21.0& 14.2& 55.1& 66.5& 39.4& 26.1& 35.1& 16.7& -& -& -& -& -\\
\multicolumn{1}{l}{BetaE}& 41.6& 11.8& 65.1& 25.7& 24.7& 55.8& 66.5& 43.9& 28.1& 40.1& 25.2& 14.3& 14.7& 11.5& 6.5& 12.4\\
\multicolumn{1}{l}{ConE}& 49.8& 14.8& 73.3& 33.8& 29.2& 64.4& 73.7& 50.9& 35.7& 55.7& 31.4& 17.9& 18.7& 12.5& 9.8& 15.1\\
\multicolumn{1}{l}{GNN-QE}& 72.8& 38.6& 88.5& 69.3& 58.7& 79.7& 83.5& 69.9& 70.4& 74.1& 61.0& 44.7& 41.7& 42.0& 30.1& 34.3\\
\multicolumn{1}{l}{CQD-CO}& 46.9& -& 89.2& 25.3& 13.4& 74.4& 78.3& 44.1& 33.2& 41.8& 21.9& -& -& -& -& -\\
\multicolumn{1}{l}{CQD-Beam}&  58.2& -& 89.2& 54.3& 28.6& 74.4& 78.3& 58.2& 67.7& 42.4& 30.9& -& -& -& -& -\\
\multicolumn{1}{l}{$\text{CQD}^{\mathcal{A}}$}& 70.4& 42.8& 89.2& 64.5& 57.9& 76.1& 79.4& 70.0& 70.6& 68.4& 57.9& 54.7& 47.1& 37.6& 35.3& 24.6\\
\multicolumn{1}{l}{QTO}& 74.0& 49.2& 89.5& 67.4& 58.8& 80.3& 83.6& 75.2& 74.0& 76.7& 61.3& 61.1& 61.2& 47.6& 48.9& 27.5\\
\multicolumn{1}{l}{CKGC}&  \textbf{80.6}& \textbf{71.0}& \textbf{89.9}& \textbf{76.5}& \textbf{72.8}& \textbf{84.6}& \textbf{88.1}& \textbf{82.1}& \textbf{80.4}& \textbf{78.7}& \textbf{72.1}& \textbf{75.0}& \textbf{74.3}& \textbf{62.6}& \textbf{70.7}& \textbf{72.5}\\
\midrule
\multicolumn{17}{c}{FB15k-237}\\
\midrule
\multicolumn{1}{l}{GQE}&  16.3& -& 35.0& 7.2& 5.3& 23.3& 34.6& 16.5& 10.7& 8.2& 5.7& -& -& -& -& -\\
\multicolumn{1}{l}{Q2B}&  20.1&  -& 40.6& 9.4& 6.8& 29.5& 42.3& 21.2& 12.6& 11.3& 7.6& -& -& -& -& -\\
\multicolumn{1}{l}{BetaE}& 20.9& 5.5& 39.0& 10.9& 10.0& 28.8& 42.5& 22.4& 12.6& 12.4& 9.7& 5.1& 7.9& 7.4& 3.5& 3.4\\
\multicolumn{1}{l}{ConE}& 23.4& 5.9& 41.8& 12.8& 11.0& 32.6& 47.3& 25.5& 14.0& 14.5& 10.8& 5.4& 8.6& 7.8& 4.0& 3.6\\
\multicolumn{1}{l}{FuzzQE}& 24.0& 7.8& 42.8& 12.9& 10.3& 33.3& 46.9& 26.9& 17.8& 14.6& 10.3& 8.5& 11.6& 7.8& 5.2& 5.8\\
\multicolumn{1}{l}{GNN-QE}& 26.8& 10.2& 42.8& 14.7& 11.8& 38.3& 54.1& 31.1& 18.9& 16.2& 13.4& 10.0& 16.8& 9.3& 7.2& 7.8\\
\multicolumn{1}{l}{CQD-CO}& 21.8& -& 46.7& 9.5& 6.3& 31.2& 40.6& 23.6& 16.0& 14.5& 8.2& -& -& -& -& -\\
\multicolumn{1}{l}{CQD-Beam}& 22.3& -& 46.7& 11.6& 8.0& 31.2& 40.6& 21.2& 18.7& 14.6& 8.4& -& -& -& -& -\\
\multicolumn{1}{l}{$\text{CQD}^{\mathcal{A}}$}& 25.7& 10.7& 46.7& 13.6& 11.4& 34.5& 48.3& 27.4& 20.9& 17.6& 11.4& 13.6& 16.8& 7.9& 8.9& 5.8\\
\multicolumn{1}{l}{QTO}& 33.5& 15.5& 49.0& 21.4& 21.2& 43.1& 56.8& 38.1& 28.0& 22.7& 21.4& 16.8& 26.7& 15.1& 13.6& 5.4\\
\multicolumn{1}{l}{CKGC}&  \textbf{34.8}& \textbf{25.3}& \textbf{49.2}& \textbf{22.3}& \textbf{22.3}& \textbf{45.1}& \textbf{60.3}& \textbf{40.3}& \textbf{29.1}& \textbf{22.9}& \textbf{22.0}& \textbf{23.9}& \textbf{37.5}& \textbf{21.0}& \textbf{23.2}& \textbf{21.2}\\
\midrule
\multicolumn{17}{c}{NELL995}\\
\midrule
\multicolumn{1}{l}{GQE}&  18.6& -& 32.8& 11.9& 9.6& 27.5& 35.2& 18.4& 14.4& 8.5& 8.8& -& -& -& -& -\\
\multicolumn{1}{l}{Q2B}&  22.9& -& 42.2& 14.0& 11.2& 33.3& 44.5& 22.4& 16.8& 11.3& 10.3& -& -& -& -& -\\
\multicolumn{1}{l}{BetaE}&  24.6& 5.9& 53.0& 13.0& 11.4& 37.6& 47.5& 24.1& 14.3& 12.2& 8.5& 5.1& 7.8& 10.0& 3.1& 3.5 \\
\multicolumn{1}{l}{ConE}& 27.2& 6.4& 53.1& 16.1& 13.9& 40.0& 50.8& 26.3& 17.5& 15.3& 11.3& 5.7& 8.1& 10.8& 3.5& 3.9\\
\multicolumn{1}{l}{FuzzQE}&  27.0& 7.8& 47.4& 17.2& 14.6& 39.5& 49.2& 26.2& 20.6& 15.3& 12.6& 7.8& 9.8& 11.1& 4.9& 5.5\\
\multicolumn{1}{l}{GNN-QE}& 28.9& 9.7& 53.3& 18.9& 14.9& 42.4& 52.5& 30.8& 18.9& 15.9& 12.6& 9.9& 14.6& 11.4& 6.3& 6.3\\
\multicolumn{1}{l}{CQD-CO}&  28.8& -& 60.4& 17.8& 12.7& 39.3& 46.6& 30.1& 22.0& 17.3& 13.2& -& -& -& -& -\\
\multicolumn{1}{l}{CQD-Beam}&  28.6& -& 60.4& 20.6& 11.6& 39.3& 46.6& 25.4& 23.9& 17.5& 12.2& -& -& -& -& -\\
\multicolumn{1}{l}{$\text{CQD}^{\mathcal{A}}$}&  32.3& 13.3& 60.4& 22.9& 16.7& 43.4& 52.6& 32.1& 26.4& 20.0& 17.0& 15.1& 18.6& 15.8& 10.7& 6.5\\
\multicolumn{1}{l}{QTO}& 32.9& 12.9& 60.7& 24.1& 21.6& 42.5& 50.6& 31.3& 26.5& 20.4& 17.9& 13.8& 17.9& 16.9& 9.9& 5.9\\
\multicolumn{1}{l}{CKGC}&  \textbf{35.3}& \textbf{19.9}& \textbf{61.4}& \textbf{25.7}& \textbf{24.1}& \textbf{45.8}& \textbf{58.8}& \textbf{33.9}& \textbf{29.2}& \textbf{20.5}& \textbf{18.7}& \textbf{19.8}& \textbf{27.5}& \textbf{20.4}& \textbf{17.4}& \textbf{14.6}\\
\bottomrule
\end{tabular}
\end{tiny}
\caption{Complex Query Answering results on FB15k, FB15k-237 and NELL995 test sets with MRR metrics. $\text{avg}_{p}$ is the average on existential positive first-order queries. $\text{avg}_{n}$ is the average on queries with negation.}
\label{table:1}
\end{center}
\end{table*}

\begin{table*}[t]
\begin{center}
\begin{tiny}
\begin{tabular}{lllllllllllllllllll}
\toprule
Models & $\text{avg}_{p}$& $\text{avg}_{n}$& 1p& 2p& 3p& 2i& 3i& pi& ip& 2u& up& 2in& 3in& inp& pin& pni\\
\midrule
\multicolumn{17}{c}{FB15k}\\
\midrule
\multicolumn{1}{l}{S12}&  78.1& 29.5& 89.9& 73.9& 68.0& 82.2& 84.9& 78.0& 77.0& 78.7& 70.0& 33.8& 34.3& 31.7& 23.7& 23.8\\
\multicolumn{1}{l}{S123}&  78.7& 45.6& 89.9& 74.0& 68.7& 83.3& 86.3& 79.4& 77.9& 78.7& 69.9& 51.4& 50.3& 45.4& 39.5& 41.3\\
\multicolumn{1}{l}{S1234}&  \textbf{80.6}& \textbf{71.0}& \textbf{89.9}& \textbf{76.5}& \textbf{72.8}& \textbf{84.6}& \textbf{88.1}& \textbf{82.1}& \textbf{80.4}& \textbf{78.7}& \textbf{72.1}& \textbf{75.0}& \textbf{74.3}& \textbf{62.6}& \textbf{70.7}& \textbf{72.5}\\
\midrule
\multicolumn{17}{c}{FB15k-237}\\
\midrule
\multicolumn{1}{l}{S12}&  32.3& 9.6& 49.2& 20.5& 19.8& 41.4& 54.8& 35.9& 26.6& 22.7& 19.9& 9.8& 16.4& 10.0& 7.1& 4.7\\
\multicolumn{1}{l}{S123}&  32.9&  15.7& 49.2& 20.9& 20.4& 42.2& 56.5& 37.1& 26.9& 22.4& 20.2& 15.8& 25.4& 14.1& 12.9& 10.2\\
\multicolumn{1}{l}{S1234}& \textbf{34.8}& \textbf{25.3}& \textbf{49.2}& \textbf{22.3}& \textbf{22.3}& \textbf{45.1}& \textbf{60.3}& \textbf{40.3}& \textbf{29.1}& \textbf{22.9}& \textbf{22.0}& \textbf{23.9}& \textbf{37.5}& \textbf{21.0}& \textbf{23.2}& \textbf{21.2}\\
\midrule
\multicolumn{17}{c}{NELL995}\\
\midrule
\multicolumn{1}{l}{S12}&  32.8& 7.3& 61.4& 23.4& 20.9& 42.5& 51.0& 31.8& 26.4& 20.4& 17.6& 7.4& 9.9& 9.9& 4.6& 4.4\\
\multicolumn{1}{l}{S123}&  33.2& 11.3& 61.4& 23.8& 21.4& 43.3& 52.5& 32.0& 26.7& 20.4& 17.8& 12.2& 16.6& 13.5& 8.0& 6.3 \\
\multicolumn{1}{l}{S1234}&  \textbf{35.3}& \textbf{19.9}& \textbf{61.4}& \textbf{25.7}& \textbf{24.1}& \textbf{45.8}& \textbf{58.8}& \textbf{33.9}& \textbf{29.2}& \textbf{20.5}& \textbf{18.7}& \textbf{19.8}& \textbf{27.5}& \textbf{20.4}& \textbf{17.4}& \textbf{14.6}\\
\bottomrule
\end{tabular}
\end{tiny}
\caption{Complex Query Answering results on FB15k, FB15k-237 and NELL995 test sets with different settings.}
\label{table:2}
\end{center}
\end{table*}

\section{Experiments}

\subsection{Experimental Settings}
\label{section:5.1}
\paragraph{Datasets}
We evaluate our method on three popular knowledge graph datasets, including FB15k \cite{bordes2013translating}, FB15k-237 \cite{toutanova2015representing}, NELL995 \cite{xiong2017deeppath}. We use the standard FOL queries generated in BetaE \cite{ren2020beta}, consisting of 9 types of existential positive first-order queries (1p/2p/3p/2i/3i/pi/ip/2u/up) and 5 types of queries with negation (2in/3in/inp/pin/pni). Specifically, "p", "i", "u", and "n" stand for "projection", "intersection", "union", and ‘negation’ in the query structure, respectively. The query types are shown in Figure \ref{figure:2}. During adaptation, we only use queries of type 1p, 2i, 3i, 2in and 3in of the training dataset to reduce computation. 

\paragraph{Baselines}
We compare our method with state-of-the-art methods on complex query answering, query embedding methods, including GQE \cite{hamilton2018embedding}, Q2B \cite{ren2020query2box}, BetaE \cite{ren2020beta}), ConE \cite{zhang2021cone}, FuzzQE \cite{chen2022fuzzy}, GNN-QE \cite{zhu2022neural}, complex query decomposition methods, including CQD-CO \cite{arakelyan2020complex}, CQD-beam \cite{arakelyan2020complex} CQD \cite{arakelyan2023adapting} and QTO \cite{bai2023answering}.

\paragraph{Evaluation Metrics}
We adopt the evaluation framework introduced in BetaE \cite{ren2020beta}, which involves categorizing the answers to each complex query into two distinct groups: easy answers and hard answers. For validation/test queries, easy answers refer to entities that can be reached by edges in training/validation graph, while hard answers are those that can only be inferred by predicting missing edges in the validation/test graph. We evaluate the method on complex queries by calculating the rank for each hard answer against non-answers and computing the mean reciprocal rank (MRR) \cite{bordes2013translating}.

\paragraph{Implementation Details}
For pre-train, our method can be incorporated with any KGC model. We select a state-of-the-art KGC model, ComplEx \cite{trouillon2017knowledge} trained with N3 regularization \cite{lacroix2018canonical} and auxiliary relation prediction task \cite{chen2021relation}. We choose the hyper-parameters with the best MRR on the validation set. 

For training of CLQA, we train at most 5 epochs. For testing of CLQA, we pre-compute the KG tensor $\bm{X}$ to make our method achieve higher efficiency compared to previous query embedding methods \cite{bai2023answering}. To save the memory usage of $\bm{X}$, we select an appropriate $\epsilon$ such that after filtering all entries that less than $\epsilon$ can be stored on a single GPU.

\begin{table*}
\begin{center}
\begin{small}
\begin{tabular}{lllllllll}
\toprule
$\epsilon$ & 0.05& 0.01& 0.005& 0.001& 0.0005& 0.0001&  0.00005& 0.00001 \\
\midrule
Memory Usage &  22M& 60M& 103M& 383M& 676M& 2.77G& 5.99G& Out of Memory\\
Sparsity Level &  99.999$\%$& 99.997$\%$& 99.995$\%$& 99.980$\%$& 99.965$\%$& 99.851$\%$& 99.678$\%$&  -\\
Running Time & 290s& 311s&  314s&  348s& 373s& 482s& 558s& -\\
$\text{avg}_{p}$&  27.9& 32.0& 32.9& 34.0& 34.3& 34.7& 34.8& -\\
$\text{avg}_{n}$ &  24.5& 25.2& 25.3& 25.3& 25.3& 25.3& 25.3& -\\
\bottomrule
\end{tabular}
\end{small}
\caption{Effect of $\epsilon$ on memory usage, sparsity level, and $\text{avg}_{p}$ and $\text{avg}_{n}$. The experiments are conducted on FB15K-237 datasets. The time is the running time on one NVIDIA A40 GPU.}
\label{table:3}
\end{center}
\end{table*}

\subsection{Results}
\label{section:5.2}
See Table \ref{table:1} for the results. $\text{avg}_{p}$ is the average on existential positive first-order queries. $\text{avg}_{n}$ is the average on queries with negation. GQE, Q2B, CKGCO, and CQD-Beam do not support queries with negation, so the corresponding entries are empty. We observe that our model significantly outperforms baseline methods across all datasets. Our model yields a relative gain of 6.8$\%$ and 52.3$\%$ on $\text{avg}_{p}$ and $\text{avg}_{n}$ compared to previous state-of-the-art model QTO. This shows that our method has better reasoning skills and superior adaptability when tackling complex query answering tasks. We attribute this improvement to our calibration method, which can adapt the KGC models to answering complex queries very well.

The adaptation function of CKGC is simple, making the model  quickly converge during the adaptation process. We next show the training time. The running time for FB15k dataset is an average of 67 seconds per epoch, the running time for FB15k -237 dataset is an average of 32 seconds per epoch, the  running time for NELL995 dataset is an average of 83 seconds per epoch. All models can converge within 5 epochs, resulting in a very short training time. In contrast, previous embedding-based models, such as BetaE, ConE, FuzzQE, GNN-QE and so on, often require hundreds of epochs to converge. For instance, BetaE takes 105 epochs to converge and has an average training running time of 63 seconds per epoch on the FB15k-237 dataset. Compared to our model (32 seconds x 2 epoch = 64 seconds), these models require significantly more runtime for training. The running time is the running time on one NVIDIA A40 GPU.

\subsection{Ablation Studies}
\label{section:5.3}
As stated in Section \ref{section:4.2}, we obtain a calibrated knowledge graph completion model by executing four steps. To analyze the impact of each step, we obtain calibrated models by performing only a few steps. We denote the calibrated model with steps 1 and 2 as S12, the model with steps 1, step 2, and step 3 as S123, and the model with steps 1, step 2, step 3, and step 4 as S1234. S12 primarily serves as a baseline. S123 is mainly used to study the impact of adaptation (step 3). S1234 is mainly used to study the impact of calibrating the predicted values corresponding to true triplets (step 4).

See Table \ref{table:2} for the results. S123 has an average relative improvement of 1.3$\%$ on $\text{avg}_{p}$ metric compared to S12, and S1234 has an average relative improvement of 4.8$\%$ on $\text{avg}_{p}$ metric compared to S123. S123 has an average relative improvement of 64.8$\%$ on $\text{avg}_{n}$ metric compared to S12, and S1234 has an average relative improvement of 64.3$\%$ on $\text{avg}_{n}$ metric compared to S123.

The results show that the $\text{avg}_{n}$ metric has been significantly improved. Since step 3 calibrates both predicted values corresponding to true triplets and predicted values corresponding to false triplets, we cannot determine from this step whether the main improvement on $\text{avg}_{n}$ comes from the calibration for true triplets or false triplets. Since step 4 only changes the predicted values corresponding to the true triplets, this indicates that the significant improvement on $\text{avg}_{n}$ is primarily due to the more accurate calibration of the predicted values corresponding to the true triplets.


\subsection{Hyper-parameters Analysis}
We analyze the experimental results of our method with respect to the hyper-parameter $\epsilon$. We study the changes in memory usage, sparsity level (the ratio of zero entries in the KG tensor $\bm{X}$), $\text{avg}_{p}$ and $\text{avg}_{n}$ as $\epsilon$ decreases. 

We show the results in Table \ref{table:3}, which show that as $\epsilon$ decreases, memory usage increases, sparsity level decreases, running time increases, $\text{avg}_{p}$ increases, and $\text{avg}_{n}$ increases. Thus, we can select a proper $\epsilon$ to balance the computation or memory usage and model performance.

\section{Conclusion}
\label{section:6}
CLQA is one of the crucial tasks associated with KGs. In this paper, we introduce CKGC, a calibration method developed to enhance the adaptability of KGC models for CLQA. Through experimental analysis, we illustrate that the implementation of CKGC leads to a substantial improvement in model performance for the CLQA task. Therefore, the calibration of KGC models holds significant importance for the optimization of CLQA models. Moving forward, we encourage the exploration and proposal of additional calibration methods to further enhance the performance of CLQA.

\section*{Limitations}
One limitation of our method is the space complexity and time complexity of projection operation. We use an approach to make the predicted tensor $\bm{X}$ sparse. This approach can make the model performance decreases as shown in Table \ref{table:3}. We plan to explore methods that can reduce the computational complexity and memory usage of the method while maintaining good performance.

\section*{Acknowledgements}
This work is supported by the National Key Research and Development Program of China (2020AAA0106000), the National Natural Science Foundation of China (U19A2079), and the CCCD Key Lab of Ministry of Culture and Tourism.

\bibliography{acl_latex}

\appendix
\onecolumn
\section{Appendix}
\label{sec:appendix}
\subsection{Experimental Details}
\paragraph{Datasets}
The detailed statistics for each dataset can be seen in Table \ref{table:4}.
\begin{table*}[ht]
\begin{center}
\begin{small}
\begin{tabular}{lllll}
\toprule
Split & Query Types& FB15k& FB15k-237& NELL995 \\
\midrule
\multirow{2}{*}{Train}&  1p,2p,3p,2i,3i& 273,710& 149,689& 107,982\\
     &  2in, 3in, inp, pin, pni& 27,371& 14,968& 10,798\\
\midrule
\multirow{2}{*}{Validation}&  1p& 59,078& 20,094& 16,910\\
   &       Others& 8,000& 5,000& 4,000\\
\midrule
\multirow{2}{*}{Test}&   1p& 66,990& 22,804& 17,021\\
                    & Others& 8,000& 5,000& 4,000\\
\bottomrule
\end{tabular}
\end{small}
\caption{Detailed statistics on the different types of query
structures in FB15K, FB15K-237, and NELL995 datasets.}
\label{table:4}
\end{center}
\end{table*}

\paragraph{Hyper-parameters Settings}
Table \ref{table:5} shows the best hyper-parameters we searched for pre-training KGC models. Table \ref{table:6} shows the best hyper-parameters we searched for CLQA.

\begin{table*}[h]
\begin{center}
\begin{tabular}{lllllll}
\toprule
Dataset & Dimension & Epoch &Batch Size &Learning Rate & $\lambda_{1}$& $\lambda_{2}$\\
\midrule
FB15k     &2000 &200 &1000 &0.01 &0.0625 & 0.005\\
FB15k-237 &2000 &200 &1000 &0.1  &4      &0.05\\
NELL995   &2000 &200 &1000 &0.1  &0.0625 &0.05\\
\bottomrule
\end{tabular}
\caption{The hyper-parameters settings for pre-training KGC models, $\lambda_{1}$ is the coefficient for relation prediction loss, and $\lambda_{2}$ is the coefficient for regularization.}
\label{table:5}
\end{center}
\end{table*}

\begin{table*}[h]
\begin{center}
\begin{tabular}{lllllll}
\toprule
Datset &Dimension &Epoch &Batch Size &Learning Rate &$\alpha$ &$\epsilon$\\
\midrule
FB15k     &2000 &5 &1000 &0.001 &0.1 &0.0005\\
FB15k-237 &2000 &5 &1000 &0.001 &0.1 &0.00005\\
NELL995   &2000 &5 &1000 &0.001 &0.1 &0.00001\\
\bottomrule
\end{tabular}
\caption{The hyper-parameters settings for CLQA.}
\label{table:6}
\end{center}
\end{table*}

\paragraph{Different KGC models}
Our proposed method CKGC is effective for any KGC model because it relies solely on the results predicted by the KGC model. Although any KGC model is permissible, it is preferable to select a KGC model with strong performance. In Table \ref{table:7}, we show the results of our method with different KGC models. We select the DistMult \citep{yang2014embedding}, CP \citep{lacroix2018canonical}, SimplE \citep{kazemi2018simple}, and ComplEx \citep{trouillon2017knowledge} KGC models. The results demonstrate that Step 3 (the adaptation step) can indeed enhance performance across various KGC models.

\begin{table*}[h]
\begin{center}
\begin{tabular}{llllll}
\toprule
Models &1p &$\text{avg}_{p}$(S12) &$\text{avg}_{p}$(S123) &$\text{avg}_{n}$(S12) &$\text{avg}_{n}$(S123)\\
\midrule
DistMult  &47.6	&30.2	&30.8	&9.1	&14.7\\
CP         &46.5	&29.4	&30.1	&8.7	&13.8\\
SimplE   &46.3	&29.0	&29.5	&8.6	&13.8\\
ComplEx &49.2	&32.4	&32.9	&9.7	&15.2\\
\bottomrule
\end{tabular}
\caption{The performance of complex logical quereis answering models with different KGC model, where 1p is the KGC results, $\text{avg}_{p}$ is the average results of queries without negation operator, $\text{avg}_{n}$ is the average results of queries with negation operator. S12 is the model using direct KGC scores. S123 is the model with adaptation.}
\label{table:7}
\end{center}
\end{table*}
\end{document}